# Cost-sensitive Semi-supervised Classification for Fraud Applications

Sulaf Elshaar and Samira Sadaoui*

Department of Computer Science University
of Regina, Regina, Canada 3737 Wascana
Parkway, SK S4S 0A2
{elshaars,Samira.Sadaoui}@uregina.ca

**Abstract**. This research explores Cost-Sensitive Learning (CSL) in the fraud detection domain to decrease the fraud class's incorrect predictions and increase its accuracy. Notably, we concentrate on shill bidding fraud that is challenging to detect because the behavior of shill and legitimate bidders are similar. We investigate CSL within the Semi-Supervised Classification (SSC) framework to address the scarcity of labeled fraud data. Our paper is the first attempt to integrate CSL with SSC for fraud detection. We adopt a meta-CSL approach to manage the costs of misclassification errors, while SSC algorithms are trained with imbalanced data. Using an actual shill bidding dataset, we assess the performance of several hybrid models of CSL and SSC and then compare their misclassification error and accuracy rates statistically. The most efficient CSL+SSC model was able to detect 99% of fraudsters and with the lowest total cost.

**Keywords**: Cost-sensitive learning, MetaCost, Cost matrix, Semi-supervised classification, Misclassification errors, Imbalanced data, Fraud detection.

## 1 Introduction

### 1.1 Problem and Motivations

Even though the auction industry is a lucrative marketplace, e-auction sites are, however, attractive to dishonest moneymakers due to the anonymity of bidders, flexibility of bidding, and cheap auction services [1]. In 2015, the Internet Crime Complaint Center reported 21,510 auction fraud complaints and a monetary loss of $19 million [12]. The illicit activities can occur before the auctions take place (ex. misrepresentation of items), during the bidding period (ex. shill bid- ding and bid shielding), and after the auctions are completed (ex. no delivery of items and fee stacking). Among the auction scams, Shill Bidding (SB) is considered the most difficult fraud to detect due to its similarity to the normal bidding behaviour. Consequently, the SB fraud goes undetected by the victims. Via alternate accounts, shill bidders compete on behalf of a seller (the auction owner) by elevating the item price without being detected. SB is still plaguing

the auction sector, as shown by several lawsuits that have been filed against dishonest sellers because SB fraud led to substantial financial losses for honest consumers [1]. SB detection is a challenging problem to address due to the following aspects: 1) thousands of auctions are held every day in auction companies, like eBay and TradeMe, 2) auctions may involve a large number of bids and bidders, 3) auctions may have long biding duration, like seven or ten days, and

4) SB identification must be made in real-time to avoid financial losses for buyers. Therefore, we adopt Machine Learning (ML) to tackle these real-life fraud scenarios. Nevertheless, we are confronted with three significant classification problems:

- Unavailability of labeled SB Data: Annotating multi-dimensional SB data is a challenging operation. Generally speaking, labeling training data is carried out by the experts of the application domain, sometimes with the help of ML techniques [1]. Still, this operation is very time-consuming.
- Absence of Misclassification Costs: Classical ML algorithms do not take into account the costs of misclassification errors and treat errors of all the classes equally. This behaviour is not appropriate in fraud detection applications where the incorrectly predicted fraud data should possess a penalty. The latter should be the highest one as the fraud data is the target for investigation.
- Presence of Class Imbalance. Fraud datasets are imbalanced. The skewed class distribution degrades the accuracy of ML algorithms [1]. Additionally, the fraud class, which is the most significant output, is misrepresented since the learning methods are influenced by the majority (normal) class.

In the previous paper [9], we addressed the labeled data scarcity problem using Semi-Supervised Classification (SSC) algorithms because they require only a few labeled data to be trained. Hence, we were able to check the ground truth of the few annotated SB data. Moreover, we empirically demonstrated that having a few annotated data during the training stage returned a satisfactory performance. We also determined the optimal amount of labeled SB data that leads to the highest accuracy. Besides, SSC can outperform supervised classification, as shown in [8]. However, SSC algorithms do not consider the costs of the misclassification errors of the two classes (Normal and Fraud). In fraud detection, this lack means that there is no difference between misclassifying a legitimate activity (normal bidding behaviour) and misclassifying a fraudulent activity (shill bidding). Nevertheless, we know that the risk of predicting a shill bidder as a normal bidder is more serious than the opposite case. Therefore, it becomes essential to take into account the misclassification costs for our fraud classification problem.

### 1.2   Contributions

In this present study, we explore Cost-Sensitive Learning (CSL) to manage both the misclassification costs and imbalanced data on the one hand, and reduce the incorrect predictions of the two classes on the other hand. In this case, we can assign a higher penalty for the fraud data that went undetected (false negatives). Indeed, we are more concerned about detecting fraudulent activities, so that we can take action against auctions infected by SB by canceling the auction before processing the payment of the item and suspending the accounts of shill bidders. We incorporate CSL into the Semi-Supervised Classification (SCC) framework. However, an SSC algorithm utilizes one or more baseline supervised classifiers that require data to be balanced. CSL addresses the skewed class imbalance at the algorithm level, i.e., without modifying the training datasets. Over- and under-sampling methods can also tackle imbalance data [1,9]. When adjusting the class distribution, over-sampling adds synthetic data to the minority class, but these data do not represent actual observations. Under-sampling method deletes some data from the majority class, which may discard essential data for the learning task. Hence, adopting CSL can be beneficial in handling the imbalanced learning problem since it employs only real bidding behaviour. We use a real SB dataset for which a small subset has been already labeled and evaluated in [9].

In our work, we take advantage of a meta-CSL approach, called MetaCost, to train semi-supervised classifiers with few labeled data that are imbalanced. We select this approach for several reasons: 1) it can be used by any type of classification algorithm, 2) it is easy to combine with a SSC algorithm, and 3) it uses ensemble learning to achieve much stronger performance, especially for unstable classifiers. With MetaCost, SSC algorithms will be able to consider the costs of misclassification errors of both classes while learning from the SB dataset. For this purpose, we define a cost matrix specifically for our SB detection problem.

We develop multiple hybrid classification models of CSL and SCC based on the cost matrix. More precisely, by varying the cost penalties of the fraud class, the most important class as it is the target for investigation, we assess and compare the accuracy and misclassification error rates of several CSL+SSC models using statistical testing. In this present study, we employ two different SSC collective packages: Chopper and Yatsi. Since in [9], CollectiveIBK returned an average performance, so in this present paper, we consider a new approach called Yatsi. We keep Chopper as it produced a very good performance. More- over, we also show that the CSL+SSC model outperforms the non cost-sensitive SSC model developed in [9] with the same SB dataset. This research is the first attempt to integrate CSL to the SSC environment in the fraud detection domain. Besides, to the best of our knowledge, we found only one recent paper that merged CSL with SSC but outside the fraud detection field, as discussed in the related work section.

We organize our paper as follows. In Section 2, we examine recent studies on CSL in the fraud detection domain and one study combining CSL and SSC. In Section 3, we describe the SB training dataset developed from commercial auctions and bidder history. In Section 4, we specify the cost matrix and describe the MetaCost method for our fraud detection application. In Section 5, we discuss multiple SSC algorithms based on two different approaches, Yatsi and Chopper, as well as the hyper-parameter tuning of both CSL and SSC methods. In Section 6, we conduct an experimental evaluation and comparison of several CSL+SSC models trained on a few labeled data that are imbalanced. In Section 7, we present essential findings of our work as well as some future research directions.

## 2   Related Work

This section reviews representative studies on cost-sensitive learning, specifically in the fraud detection domain. This review will allow us to examine the costs assigned to the incorrect prediction of the fraud class. Nevertheless, prior works have almost exclusively focused on detecting fraud in credit card transactions. We believe this limitation is due to the availability of training datasets, on the one hand, and complaints of victims who reported the fraud on the other hand. Moreover, the ability to label transactions as normal or fraud provided valuable support to the research. Several studies examined SSC for fraud detection, but we found only one recent paper that combined CSL and SSC to the best of our knowledge.

The study [3] investigated the ensemble of CSL and Bayesian network to detect credit card fraud. The annotated training dataset was provided by "UOL PagSeguro", a Brazilian online payment company. For the CSL task, the authors adopted two methods to deal with imbalanced data: instance re-weighing and class probability threshold. To assess the fraud model's performance, they considered two metrics only: F1-score and the cost named "economic efficiency". However, due to data privacy, they did not mention the actual values of costs. Instead, they provided an equation to compute the costs from the transactions, which is used in the current system of PagSeguro company. The class probability threshold led to the best accuracy. Moreover, the authors stated that a model with a high accuracy does not necessarily have a low cost.

Another research [17] also detected credit card fraud based on CSL. First, the authors collected labeled data from an anonymous bank from 2012 to 2013. Additionally, they considered the deviation from the normal behavior of customers as a sign of fraud. Next, they defined an equation to calculate the costs from the transactions, but they did not provide the cost amounts in the experiments. For the classification task, they employed Random Forest (RF) and then the hybrid version of CSL and RF. The findings demonstrate that utilizing CSL reduced the misclassification errors of the fraud class by 23%.

The paper [10] incorporated CSL to Neural Networks to detect credit card fraud. The dataset was supplied by the company "BBVA Data & Analytics," which consists of anonymous card transactions for 2014 and 2015. Besides, the company also provided the fraud claims, which made it easier to label customers as normal or fraud. The authors removed a large number of transactions because the dataset was highly imbalanced, with a ratio of fraudulent to normal data equals to 1:5000. They considered the cost as the amount of money associated with the fraud activities detected by the model. Based on the monetary cost, the experiment showed that the fraud model achieved similar accuracy that was previously attained by other costly models. However, the values of the cost were not given in this study, and it was not clear how the CSL matrix was employed in the supervised framework.

The study [11] examined how different values of the cost of False Negatives (i.e., fraud wrongly classified as normal) can affect the performance of CSL models. The authors trained the Bayes Minimum Risk algorithm along with several base CSL classifiers using different values of FN costs. They trained the CSL classifiers with a credit card dataset published in 2009 by the UCSD repository. First, they adopted the average number of transactions as the value of the FN cost, and then used numerous random values that are lower or larger than the average value. The results showed that CSL models produced different results when using different costs of FNs. The lower the FN cost, the better the model performance.

In [18], the authors utilized a cost-sensitive Decision Tree method to minimize the misclassification errors when detecting fraud in credit card data. The labeled dataset was provided by an anonymous bank. The authors varied the costs of FNs based on the available limit of the credit card transactions. The experiments demonstrated that the hybrid model CSL+Decision Tree outper- formed traditional classifiers, such as Artificial Neural Networks, Decision Trees and SVM, in terms of accuracy, true positive rate, and misclassification errors.

Very recently, [23] developed an ensemble GMDH Neural Network method based on CSL and SSC to identify customers with good or bad credit. This scoring can assist financial companies to make decisions regarding customer loan approval. The authors assessed the proposed method with an old public labeled credit scoring dataset (from 2009 to 2011). Intending to fill the gap in the literature, they used CSL to handle imbalanced data and SSC because, in many cases, the labels were not provided. The accuracy results showed that the developed model was superior to standard SSC models, such as CoBag, Semi-bagging, and Tri-training. The experiments also proved that fewer labeled samples led to a better scoring performance.

## 3  Fraud Dataset Overview

We developed a reliable SB dataset using a large collection of commercial auctions of eBay and their bidder history too [7] (see Table 1). We rigorously preprocessed the two crawled datasets, auctions and bidders. Then, we implemented a collection of nine SB strategies exposed in Table 1. For more details about the fraud patterns and their measurement algorithms, consult the article [7].

6       S. Elshaar et al.Subsequently, for each bidder in each auction, we evaluated the metric of each SB pattern. This measurement task resulted in an SB dataset consisting of 9291 samples (after removing outliers). An SB sample denotes the behaviour of a bidder in an individual auction, which is Normal or Fraud. It is a vector of eleven elements: Bidder ID, Auction ID, and the nine SB patterns. With this granularity, we can act against each auction infected by fraud to avoid a monetary loss for the winning bidder.

In a subsequent work [9], we appropriately labeled a small portion of the SB dataset to conduct the semi-supervised classification task. For this purpose, we first combined two data clustering techniques, X-means and Hierarchical clustering, to produce clusters of bidders of high quality. Then, we proposed a new approach to detect fraudulent activities or anomalies in each cluster based on the bidders´ B scores in that cluster and the Three Sigma Rule. Lastly, we experimented to determine the minimal sufficient amount of labeled data statistically to achieve the highest accuracy [9]. In Table 1, we can observe that the SB labeled subset is imbalanced with a ratio of Normal to Fraud samples equals to 5:1.

Table 1. Fraud Dataset and its Labeled Subset

| Number of Auctions | 1399 | |
|---|---|---|
| Number of Bidders | 1100 | |
| Bidding Duration | 1, 3, 5, 7 and 10 days | |
| Number of Samples | 9291 | |
| Fraud Predictors | - Bidder Tendency<br>- Bidding Ratio<br>- Last Bidding<br>- Auction Bids<br>- Starting Price<br>- Early Bidding<br>- Winning Ratio<br>- Buyer Rating Based on Items<br>- Bid Retraction | |
| Labeled Subset | Normal | Fraud |
| (total: 945) | 791 | 154 |
| Unlabeled Subset | 8346 | |

## 4  Cost-Sensitive Learning Framework

### 4.1  Cost Matrix for SB Fraud

Real-life detection applications can be endowed with different types of costs that can be utilized to improve further their prediction outcome. The costs are mostly financial, as the cost of hiring experts, or using specific devices, or conducting additional tests. However, the cost can also be non-financial but paramount, such as the cost of identifying the disease carrier as not carrying it, or the cost of classifying a fraudster as a genuine bidder. In our specific problem, we cannot estimate the monetary cost because we do not know objectively the loss that occurs when the classification result is erroneous. However, we know for sure that the risk of classifying a shill bidder as a normal bidder is higher than the risk of classifying a normal bidder as a shill bidder. So, it becomes essential to consider the penalties for the wrong predictions in our SB detection application.

In a two-class classification problem, a prediction can be one of four outcomes: True Positive (TP) and True Negative (TN) are the correct predictions whereas False Positive (FP) and False Negative (FN) are the incorrect predictions. In our application, FPs represent honest bidders misclassified as shill bidders, and FNs shill bidders mislabeled as honest bidders. CSL-based models utilize a cost matrix to assign relevant costs for the misclassification errors. Generally speaking, if the costs are known apriori or can be provided by the experts of the application domain, we can assign different costs for the incorrect predictions and different benefits for the correct predictions. In this case, the cost matrix can be pro- vided directly to the cost-sensitive classifiers. Nevertheless, in our classification problem, the costs are unknown. We are more interested in detecting fraudulent bidding behaviour so that we can take action against infected auctions. There- fore, we put more penalty on the fraud samples that went undetected (i.e., FNs).

As presented in Table 2, we set the cost matrix to 2x2 because we have two target classes. Since we are not studying the profits of the correct predictions, we assign to their penalties the default value i.e., $Cost_{TP} = Cost_{TN} = 0$. Regarding the mislabelled instances, we set the penalty of FPs to "1"; however no rules can be found in the literature for choosing the values for the FN penalty. After examining the literature, the most common costs employed in past empirical studies ranged from 1 to 10. As an example, the paper [15] used the values of 1, 2, 3, 4, 6 and 10, [5] selected 2, 5 and 10, and [14] chose 2, 3, 4, 5, 6, 7, and 8. We note that no reasons have been mentioned for choosing certain values over others.

We vary the FN costs from 2 to 5 and keep FP cost equal to 1 with the sole purpose of preventing more penalties on the SSC models when they predict instances incorrectly. First, we choose the values of 2, 3, 4 and 5 because they are the most common in previous research. Second, we do not consider values that are higher than the class imbalance ratio of 5:1 of our labeled SB subset (see Table 1). A higher value means too many penalties on the classifiers, which may lead to many errors when classifying normal bidders as fraudsters. Table 2 presents our cost matrix for the CSL+SSC models.

The goal of the CSL is to develop a classifier with the lowest total cost, which is calculated as follows [15]:

$$TotalCost = Count_{FNs} * Cost_{FN} + Count_{FPs} * Cost_{FP} \tag{1}$$

Table 2. Cost Matrix for SB Classification

|              |        | Predicted Class | |
|--------------|--------|-----------------|-------|
|              |        | Normal          | Fraud |
| Actual Class | Normal | 0               | 1     |
|              | Fraud  | 2, 3, 4, 5      | 0     |

where $Count_{FNs}$ denotes the count of FNs and $Count_{FPs}$ the count of FPs.

On another note, CSL models are capable of dealing efficiently with imbalanced data at the algorithm level. Consequently, the SSC algorithms will learn from the real bidding behavior of users since we are not altering the SB subset, unlike with data sampling techniques.

### 4.2 MetaCost Learning

Traditionally, to make a two-class classifier cost-sensitive, a CSL algorithm modifies the proportion of training samples when the misclassification errors have different penalties. This technique, called data stratification, consists of modifying the proportion of the minority class because it has the highest risk. The goal here is to minimize the prediction errors of this class [6]. Changing the proportion of the minority class can be done by either duplicating the instances or re-weighting the instances to the relative cost of errors of FNs and FPs [13]. The disadvantage of the first option is that it changes the training dataset, but our goal is to consider only the real behaviour of bidders. On the other hand, if data stratification is done by re-weighing the instances, we are forced to utilize only those algorithms that possess this capability. Thus, we prefer not to be restricted to a specific type of learning algorithm that has the ability of re-weighting data samples.

Due to the disadvantages mentioned above, we adopt a meta-CSL approach, called "MetaCost", which can make any classifier cost-sensitive but without data stratification [5]. MetaCost is appropriate for our fraud classification problem owning to the following factors:

- It can be utilized by any type of classification algorithm, supervised and semi-supervised, and algorithms that can re-weight or not the instances.
- It is the best candidate to manage the penalties of multi-class classifiers [13]. MetaCost works by wrapping the cost-minimizing concept around the classifier but without knowing its internal procedures [5].
- It uses the concept of bagging together with the costs. Bagging produces very accurate probability estimates for unstable classifiers, which are achieved by creating many bootstrapping randomly. Bootstrapping together with the cross-validation optimization method allows a classifier to be trained with a wide variety of samples.

- It employs ensemble learning in which the final prediction decision is made by several classifiers using the weighted majority voting. A fraud model based on multiple classifiers' decisions is much stronger than the best model obtained by one classifier.

We first specify the misclassification costs of both classes in MetaCost as shown in the cost matrix. We then develop four CSL+SSC models by integrating MetaCost into each classification algorithm. Each SSC algorithm employs base classifiers that are trained with the labeled SB subset (imbalanced).

## 5 Experiment Framework

### 5.1 SSC Algorithms

Regarding the SSC framework, we select two different collective packages, called "Chopper" [21] and "Yatsi" (Yet Another Two-Stage Idea) [4]. Chopper is an ensemble learner that works only for the two-class classification task. It employs a first classifier to label the testing data after being trained with the labeled subset. This classifier determines the distributions for all the testing data and then ranks the data based on the difference between two confidences (classes). The new training dataset is then supplied to a second classifier, which again determines the distributions for the remaining testing data. The Yatsi approach conducts the classification in two phases: 1) it trains the first classifier with the small portion of labeled data, 2) it employs the classifier to transform the unlabeled data into weighted data (called pre-labeled data), and lastly 3) after merging the original labeled data and pre-labeled data, it uses KNN to produce the classes of the pre-labeled data using the following approach: KNN sums the weights of the nearest neighbours of the pre-labeled data, and then label those data with the class that has the largest summation of weights.

We customize Chopper and Yatsi with the classification algorithms that are commonly adopted in the field of fraud detection, including Naive Bayes (NB), Random Forest (RF), J48 (implementation of Decision Trees C4.5) and IBK (implementation of KNN). More precisely, we develop Chopper with NB as the first classifier and RF as the second classifier. With Yatsi, we train Yatsi-J48, Yatsi-KNN, and Yatsi-NB.

### 5.2 Parameter Tuning

We optimize the hyper-parameters of the SSC models using a predefined class named "CVParameter" of Weka toolkit. This class determines the optimal values of the parameters using Cross-Validation (CV). However, we need to supply which parameters to be tuned, and their range of values. We train the SSC algorithms using 10-fold CV and perform ten runs to obtain more stable SB classifiers.

For the second classifier of Chopper, we first set the range of the two RF hyper-parameters: the Number of Iterations (NI) from 50 to 300 and the Maxi- mum Tree Depth (MTD) from 1 to 50. CVParameter returns the optimal classifier with the lowest FNR using the best parameter values: 100 for NI and 12 for MTD. We train Yatsi-KNN classifier by setting the number of Nearest Neighbours (NNs) to 5 because this value led to the lowest FNR. We choose the KDTree search algorithm to accelerate the NN search. We also assign the value of 1.0 (default value) to the weighting factor for unlabeled data. Lastly, we tune the two hyper-parameters of J48: the pruning tree confidence factor with the optimal value of 0.75, and the minimum number of data per leaf with the best value of 2.

For the ensemble learning task used by MetaCost, we assign the number of bagging iterations to 10 and the batch size to 100. These default values are the most preferred when the prediction is performed [21].

### 5.3   Performance Metrics

Since we are more concerned about detecting shill bidders, we, therefore, focus on the fraud data. Moreover, the CSL+SSC models that we develop are trained directly with the imbalanced SB subset. Hence, we choose the most relevant performance metrics for imbalanced data and correctly predicted fraudsters.

- Recall returns the ratio of shill bidders correctly and wrongly classified.

$$Recall = \frac{TP}{TP + FN} \quad (2)$$

- False Negative Rate (FNR) measures the ratio of shill bidders wrongly classified as normal bidders.

$$FNR = \frac{FN}{FN + TP} \quad (3)$$

- False Positive Rate (FPR) calculates the ratio of normal bidders wrongly classified as fraudulent.

$$FPR = \frac{FP}{FP + TN} \quad (4)$$

- Kappa Statistic (Kappa) computes the agreement between actual and predicted classes while correcting the agreement that happens by chance.

$$KappaStat = \frac{P_{observed} - P_{chance}}{1 - P_{chance}} \quad (5)$$

- Area Under the ROC Curve (AUC) informs us of how much a classifier can differentiate between the normal and fraud class. The closer the AUC value is to 1, the better is the classification model.

## 6 Evaluation and Comparison

### 6.1 Performance Results

We first report in Table 3, 4, 5 and 6 the performance results of the four SB classifiers by varying the misclassification cost of the fraud class. The results are then discussed and compared in the next sections.

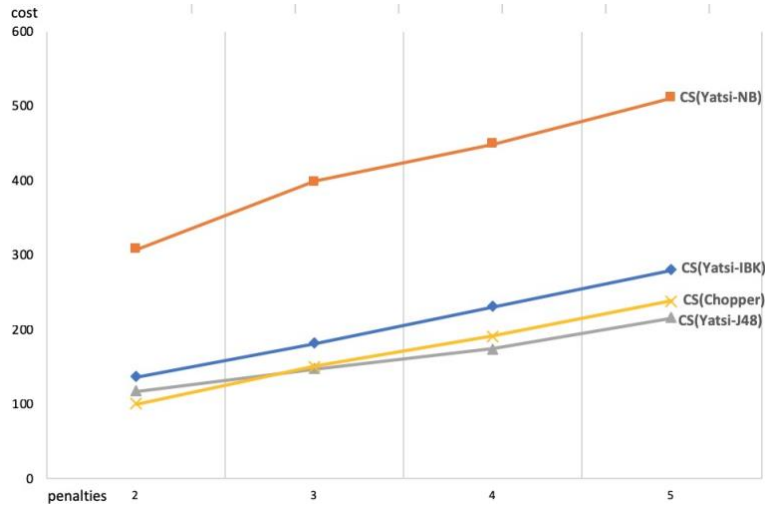

Fig. 1. Learning Curves of CSL+SSC Models with Different Penalties

### 6.2 Misclassification Errors

In terms of minimizing the misclassification error of the fraud class, the classification model with the lowest cost is the best. As presented in Table 3, when the cost penalty is 2, CSL+Chopper provided the lowest cost of 100 followed by CSL+Yatsi-KNN with 137. Regarding the other penalties of 3, 4 and 5, CSL+Yatsi-J48 returned the lowest cost of 147, 174 and 216 respectively, which is followed by CSL+Chopper with a gap of 4, 17 and 22 respectively. On the other hand, CSL+Yatsi-NB has the highest cost across all the penalties. Moreover, as illustrated in Figure 1, the learning curves of the CSL+SSC models demonstrate that the misclassification errors increase dramatically when the penalties are in- creasing. Therefore, the best performing model is CSL+Chopper with the cost penalty of 2.

Table 3. CSL+SSC Performance when Penalty is 2, "*" indicates the model is significantly worse while "**" significantly better

|        | CSL+Yatsi-KNN | CSL+Yatsi-NB | CSL+Yatsi-J48 | CSL+Chopper |
|--------|---------------|--------------|---------------|-------------|
| Kappa  | 0.66          | 0.14 *       | 0.71          | 0.76 **     |
| FNR    | 0.03          | 0.1 *        | 0.04          | 0.01        |
| FPR    | 0.36          | 0.77 *       | 0.27          | 0.28        |
| Recall | 0.97          | 0.9          | 0.96          | 0.99        |
| AUC    | 0.92          | 0.76 *       | 0.86 *        | 0.97        |
| Cost   | 137           | 308          | 177           | 100         |

Table 4. CSL+SSC Performance when Penalty is 3, "*" indicates the model is significantly worse

|        | CSL+Yatsi-KNN | CSL+Yatsi-NB | CSL+Yatsi-J48 | CSL+Chopper |
|--------|---------------|--------------|---------------|-------------|
| Kappa  | 0.65          | 0.19 *       | 0.7           | 0.75        |
| FNR    | 0.05          | 0.12 *       | 0.05          | 0.03        |
| FPR    | 0.31          | 0.69 *       | 0.25          | 0.27        |
| Recall | 0.95          | 0.88         | 0.95          | 0.97        |
| AUC    | 0.93          | 0.79 *       | 0.86 *        | 0.96        |
| Cost   | 182           | 398          | 147           | 151         |

Table 5. CSL+SSC Performance when Penalty is 4, "*" indicates the model is significantly worse while "**" significantly better

|        | CSL+Yatsi-KNN | CSL+Yatsi-NB | CSL+Yatsi-J48 | CSL+Chopper |
|--------|---------------|--------------|---------------|-------------|
| Kappa  | 0.62          | 0.27 *       | 0.67          | 0.72 **     |
| FNR    | 0.07          | 0.15 *       | 0.07          | 0.04 **     |
| FPR    | 0.29          | 0.56 *       | 0.23          | 0.25 **     |
| Recall | 0.93          | 0.85 *       | 0.93          | 0.96 **     |
| AUC    | 0.93          | 0.79 *       | 0.88 *        | 0.96        |
| Cost   | 231           | 449          | 174           | 191         |

Table 6. CSL+SSC Performance when Penalty is 5; "*" indicates the model is significantly worse while "**" significantly better

|        | CSL+Yatsi-KNN | CSL+Yatsi-NB | CSL+Yatsi-J48 | CSL+Chopper |
|--------|---------------|--------------|---------------|-------------|
| Kappa  | 0.62          | 0.28 *       | 0.66          | 0.7         |
| FNR    | 0.09          | 0.19 *       | 0.08          | 0.05 **     |
| FPR    | 0.24          | 0.48 *       | 0.22          | 0.23 **     |
| Recall | 0.91          | 0.81 *       | 0.92          | 0.95 **     |
| AUC    | 0.93          | 0.79 *       | 0.88 *        | 0.95        |
| Cost   | 280           | 511          | 216           | 238         |

### 6.3 Statistical Comparison

We also compare all the accuracy of the four classifiers based on the statistical testing T-test. The symbol "*" indicates significant worse performance while "**" significant better performance.

- When the cost penalty is 2, the accuracy of CSL+Chopper is significantly better with KappaStat because 76% of data are really correctly classified, and not by chance. There is no significant difference with the other metrics. CSL+Yatsi-NB is significantly worse across all the metrics.
- When the cost penalty is 3, we observe no significant difference between CSL+Yatsi-KNN, CSL+Yatsi-J48 and CSL+Chopper. However, CSL+Yatsi- J48 returned the worse AUC of 0.86. CSL+Yatsi-NB is again the worst across all the metrics.
- When the cost penalty is 4, the performance of CSL+Chopper is again signif- icantly better with KappaStat because it outperforms CSL+Yatsi-KNN by 10%, produced 3% less false negatives, and detected 99% of fraud. CSL+Yatsi- NB is the worst across all the metrics.
- When the cost penalty is 5, CSL+Chopper is significantly better in terms of FNR, FPR and Recall. However, there is no statistical difference between CSL+Chopper and CSL+Yatsi-KNN in terms of KappaStat and AUC. The worst AUC values were generated by CSL+Yatsi-J48 and CSL+Yatsi-NB.

In conclusion, the most performing hybrid model is CSL+Chopper, which provided the lowest total cost of 100 when penalizing the classifier with the cost of 2. We observe that 76% of the predictions are really correctly classified and not by chance, only 1% of shill bidders are undetected, and 28% of normal bidders are misclassified. Moreover, by comparing CSL+Yatsi-IBK with CSL+Chopper, we find out that both models are very accurate in classifying bidders into normal and fraudsters with a Recall of 0.97 and 0.99 respectively.

Furthermore, the classifier CSL+Chopper outperforms the regular Chopper model developed in [9] by minimizing the misclassification error of the fraud class by 14% and increasing the accuracy by 17%. Indeed, Chopper (using re-balanced data) returned a FNR of 0.15 and CSL+Chopper (using original imbalanced data) a FNR of 0.01. Chopper provided a Recall of 0.82 and CSL+Chopper a Recall of 0.99. Since these gaps are significant in the fraud detection field, we can conclude that CSL+SSC classifier is the best fit for our fraud detection problem.

## 7 Conclusion and Future Work

In our study, we successfully incorporated a meta cost-sensitive learning method into the semi-supervised classification environment to achieve three essential benefits: 1) learn with few labeled data because it is time-consuming to annotate multi-dimensional fraud data, 2) manage the costs of misclassification errors so that the fraud class, which is the target of the investigation, has the highest penalty, and 3) tackle the imbalanced learning problem at the algorithm level, so that only the real behaviour of bidders/users is considered during training. Using a real fraud dataset, we conducted an in-depth evaluation and comparison of the performance of several hybrid models of cost-sensitive and semi-supervised classifiers. CSL+Chopper (based on Naive Bayes and Random Forest) is the most performing model concerning the error minimization and accuracy maximization. This fraud model was able to detect 99% of shill bidders with the lowest total cost of 100. Also, CSL+Chopper outperforms the regular Chopper.

For future work, we plan to investigate the interactive active learning method. The latter first chooses the most relevant data and then asks the developer or expert to suggest a label for the chosen data. Our goal is to examine how semi- supervised classifiers would be affected by the selected labeled data. We are also interested in conducting a comparison of our semi-supervised classifiers with the chunk-based incremental classification method defined in [2]. The latter has also been used to tackle the scarcity of labeled data in the fraud detection domain.